\documentclass{llncs}
\usepackage[utf8]{inputenc}
\usepackage{amssymb,amsmath}
\usepackage{graphicx}
\usepackage[space]{grffile}
\usepackage[caption=false]{subfig}
\usepackage{xspace}

\newcommand{\ie}{i.e.}

\newcommand{\Real}{\mathop{\rm I\kern-.2emR}}
\newcommand{\set}[1]{\ensuremath\left\{{#1}\right\}}
\newcommand{\Neig}[0]{\ensuremath{\mathcal N}}
\newcommand{\DS}[0]{\ensuremath{X}}
\newcommand{\OS}[0]{\ensuremath{Z}}

\newcommand{\dominates}{\prec}
\newcommand{\ndominates}{\nprec}

\newcommand{\Better} { \vartriangleleft }

\newcommand{\RMNK}{\ensuremath{\rho}MNK-landscape\xspace}
\newcommand{\RMNKs}{\ensuremath{\rho}MNK-landscapes\xspace}
\newcommand{\PLSunb}{PLS\ensuremath{_\textsc{unb}}\xspace}
\newcommand{\PLShva}{PLS\ensuremath{_\textsc{hva}}\xspace}
\newcommand{\PLSmga}{PLS\ensuremath{_\textsc{mga}}\xspace}

\newcommand{\MaxMinAntSystem}{\text{$\cal MAX$-$\cal MIN$} {A}nt {S}ystem\xspace}

\usepackage[bookmarksopen]{hyperref}
\hypersetup{
   colorlinks=true,       
   linkcolor=black,          
   citecolor=black,        
   filecolor=black,      
   urlcolor=black           
}
\usepackage{algorithmic,algorithm}
\usepackage{color}

\title{Local Optimal Sets and Bounded Archiving\\on Multi-objective
  NK-Landscapes\\with Correlated Objectives}
  
\author{Manuel L{\'o}pez-Ib{\'a}{\~n}ez\inst{1} \and 
  Arnaud Liefooghe\inst{2} \and S{\'e}bastien Verel\inst{3}} 

\institute{%
IRIDIA, Universit{\'e} Libre de Bruxelles (ULB), Brussels, Belgium
\email{manuel.lopez-ibanez@ulb.ac.be}
\and
Universit\'e~Lille~1, LIFL, UMR CNRS 8022, Inria Lille-Nord Europe, France
\email{arnaud.liefooghe@univ-lille1.fr}
\and
Universit\'e du Littoral C\^ote d'Opale, LISIC, France
\email{verel@lisic.univ-littoral.fr}
}
\begin{document}
\maketitle

\begin{abstract}
  The properties of local optimal solutions in multi-objective
  combinatorial optimization problems are crucial for the
  effectiveness of local search algorithms, particularly when these
  algorithms are based on Pareto dominance.
  Such local search algorithms typically return a set of mutually
  nondominated Pareto local optimal (PLO) solutions, that is, a PLO-set.
  This paper investigates two aspects of PLO-sets by means of experiments
  with Pareto local search (PLS). First, we examine the impact of
  several problem characteristics on the properties of PLO-sets for
  multi-objective NK-landscapes with correlated objectives. In
  particular, we report that either increasing the number of
  objectives or decreasing the correlation between objectives leads to
  an exponential increment on the size of PLO-sets, whereas the
  variable correlation has only a minor effect. Second, we study the
  running time and the quality reached when using bounding archiving methods
  to limit the size of the archive handled by PLS, and thus, the maximum
  size of the PLO-set found. We argue that there is a clear
  relationship between the running time of PLS and the difficulty of a
  problem instance.
\end{abstract}

\section{Introduction}

Several state-of-the-art algorithms for multi-objective combinatorial
optimization problems (MCOPs) are based on local search. These local
search algorithms are either based on solving multiple scalarizations of
the objective function vector, or they are based on Pareto
dominance. The most successful local search algorithms combine both
approaches~\cite{DruThi2012,DubLopStu2011cor,Lust09,PaqStu06:mqap}. 
These successes explain the increasing interest on
understanding the role of local optimal solutions in the context of
MCOPs.
Previous works have focused on the properties of individual local
optimal solutions~\cite{VerLieJou2013ejor}. However,    
algorithms for MCOPs typically return not a single solution, but a set
of solutions that approximates the Pareto 
set. Thus, local search algorithms for MCOPs are typically concerned
by (Pareto) local optimal \emph{sets} (PLO-sets), that is, sets of
(Pareto) local optimal solutions where solutions are mutually
nondominated and are also local optimal with respect to the
neighborhood of the other solutions in the set~\cite{PaqSchStu07:aor}.

Experimental studies on PLO-sets have been so far
limited to the study of some of their properties for the
bi-objective traveling salesman problem, in particular the number of
solutions in each PLO-set and the connectedness of
PLO-sets~\cite{PaqChiStu2004mmo}. 
This paper extends significantly this initial work in two aspects.
First, we consider multi-objective NK-landscapes with correlated
objectives (\RMNKs)~\cite{VerLieJou2013ejor}, which allow us to
examine the effect of various problem characteristics on the
properties of PLO-sets. Second, we examine the effect of using bounded
archiving methods~\cite{KnoCor2004lnems,LopKnoLau2011emo} in order to
limit the size of the PLO-set handled by the local search. Archiving
methods are often used in both local search and evolutionary
multi-objective algorithms, and there are a few
recent works on their theoretical
properties~\cite{BriFri2011gecco,LauZen2011ejor,LopKnoLau2011emo}. Several
authors have mentioned as interesting future work the experimental
study of the PLO-sets induced by bounded archiving. To the best of our
knowledge, we present in this paper the first results of such a
study. Our conclusions not only give support to previous theoretical
results, but also give insights on how to improve local search
algorithms for particularly difficult MCOPs.
In the following, we recall the required background on problems and algorithms
in Section~\ref{sec:background}; we provide the experimental setup of our
study in Section~\ref{sec:exp}; we discuss our experimental results in
Section~\ref{sec:res}; and we conclude in the last section.

\section{Background}
\label{sec:background}

\subsubsection{Multi-objective Combinatorial Optimization.}

A \emph{multi-objective combinatorial optimization problem} (MCOP) is defined
by an objective function vector $f=(f_1,\dotsc, f_m)$ with $m \geq 2$
objective functions, and a discrete set $\DS$ of feasible solutions in
the \emph{solution space}.
Let $\OS = f(\DS) \subseteq \mathbb{R}^m$ be the set of feasible
outcome vectors in the \emph{objective space}.  Each solution $x
\in \DS$ is assigned an objective vector $z \in \OS$ on the basis of
the multidimensional function vector $f \colon \DS \to \OS$ such
that $z = f(x)$.
When all objectives are to be maximized,
a solution $x$ weakly dominates another solution $x'$ if
$\forall i \in \set{1,\dotsc,m}$, $f_i(x) \geq f_i(x')$.
If, in addition, $\exists i \in \set{1,\dotsc,m} $ such that
$f_i(x) > f_i(x')$, then we say that $x$ dominates $x'$
($x \dominates x'$).
When $x \ndominates x' \land x' \ndominates x$, we say that $x$ and
$x'$ are mutually nondominated.
A solution $x^\star \in \DS$ is \emph{Pareto optimal}
if $\nexists x \in \DS$ such that $x \dominates x^\star$. 
The set of all Pareto optimal solutions is the \emph{Pareto set}. 
Its mapping in the objective space is the \emph{Pareto front}. 
One of the goals in multi-objective optimization is to identify the Pareto set, or a good approximation of~it.

\subsubsection{Pareto Local Optimal Sets.}

Set-based local search algorithms for MCOPs generally combine the use
of a neighborhood operator with the management of an archive of
mutually nondominated solutions found so far.
A \emph{neighborhood operator} is a mapping function $\Neig \colon \DS \to 2^\DS$ 
that assigns a set of solutions $\Neig(x) \subset X$ to any solution $x \in X$.
$\mathcal{N}(x)$ is called the \emph{neighborhood} of $x$,  and a solution $x^\prime \in \Neig(x)$ is called a \emph{neighbor} of $x$.
A solution $x \in \DS$ is a \emph{Pareto local optimum (PLO)} with
respect to a neighborhood structure~$\Neig$ if there is no neighbor
$x^\prime \in \Neig(x)$ such that $x' \dominates
x$~\cite{PaqSchStu07:aor}. 
A set $S \subseteq
\DS$ is a \emph{Pareto local optimal set (PLO-set)} with respect to
$\Neig$ if, and only if, it contains only PLO-solutions with respect
to $\Neig$ and all solutions are mutually nondominated~\cite{PaqSchStu07:aor}.
In addition, a set $S \subseteq \DS$ is a \emph{maximal PLO-set} with
respect to $\Neig$ if, and only if, $\forall s' \in \Neig(S)$,
$\exists s \in S$ such that $s \dominates s' \lor f(s) = f(s')$, where
$\Neig(S) = \bigcup_{s \in S} \Neig(s)$~\cite{PaqSchStu07:aor}. In
other words, any neighbor of any solution in a maximal PLO-set is
weakly dominated by a solution in the set.
\subsubsection{Pareto Local Search.}

A typical example of a multi-objective local search algorithm is
Pareto Local Search~(PLS)~\cite{PaqSchStu07:aor}.
PLS is an extension of the
conventional hill-climbing algorithm to the multi-objective case.  An
archive of nondominated solutions is initialized with at least one
solution.
At each iteration, one solution is chosen at random from the archive
and all its neighbors are evaluated and compared against the
archive. Each neighbor is added to the archive, and marked as
\emph{unvisited}, if it is not dominated by any other solution in the
archive. Moreover, solutions in the archive dominated by this neighbor
are removed. Once all neighbors have been evaluated, the current
solution is marked as \emph{visited}.  The algorithm stops once all
solutions from the archive are marked as \emph{visited}.  

PLS is known to be a well-performing algorithm, either as a
stand-alone approach or as a hybrid component, for many
MCOPs~\cite{DruThi2012,DubLopStu2011cor,Lust09,PaqStu06:mqap}. %
Moreover, independently of the initial archive, PLS always terminates
and returns a maximal PLO-set~\cite{PaqSchStu07:aor}.
However, the archive of (unvisited) solutions may grow exponentially
with respect to the instance size and, in that case, PLS may require
an exponential number of iterations. In such a situation, it would be
more interesting to bound the size of the archive in order to prevent
an exponential grow, but still return a (perhaps non-maximal)
PLO-set. 

Given an archive $A$ and a maximum size $\mu$, a bounded archiving
method, or \emph{archiver} for short, will return a new archive $A'
\subseteq A$ such that $|A'| \leq
\mu$~\cite{KnoCor2004lnems,LopKnoLau2011emo}.  We can use an archiving
method in PLS such that whenever a new solution $x'$ is added to the archive $A$
and $|A \cup \{x'\}| = \mu + 1$, then the archiving method will select
one solution to be removed.
This is equivalent to the $\mu+\lambda$ strategy~\cite{BriFri2011gecco}, with $\lambda=1$.
The various archiving methods differ on how the solution to be removed
is selected. Here we focus on two archiving methods, hypervolume
archiver (HVA)~\cite{Knowles2002PhD} and multi-level grid archiver
(MGA)~\cite{LauZen2011ejor}, which are the only known archiving
methods belonging to the class with the most desirable convergence
properties~\cite{LopKnoLau2011emo}. Two of these properties are: 
\emph{(i)} accepting solutions outside the objective space region dominating
the current archive (\emph{diversifies}) and \emph{(ii)} a subsequent
archive cannot be worse in terms of Pareto dominance than an earlier
archive (\emph{$\Better$-monotone}).

When PLS uses either HVA or MGA as its archiving method, then a
run of PLS will stop at a PLO-set, but not necessarily at a maximal
PLO-set. Indeed, there may exist a solution $s' \in \Neig(A)$, such
that $\nexists s \in A$ with $s \dominates s'$, but the archiving
method chose to discard $s'$ in order to maintain $|A| \leq \mu$.
Therefore, it is expected that when using an archiving method,
PLS will converge faster but to a possibly worse PLO-set. Moreover,
since the decision of which solutions are discarded are fundamentally
different for HVA and MGA, we would expect that each method may
converge to disjoint sets of PLO-sets. %
In the experiments presented here, we study the properties of the
PLO-sets returned by the classical PLS (using an unbounded archive)
and when PLS uses either HVA or MGA to bound the archive size.

\subsubsection{Multi-Objective NK-landscapes with Correlated Objectives.}

We study the effect of various characteristics of MCOPs on the
properties of PLO-sets by means of \RMNKs, which are artificial multi-objective
multimodal problems with objective correlation~\cite{VerLieJou2013ejor}.
They extend both single-objective NK-landscapes~\cite{Kau1993order} and
multi-objective NK-landscapes with independent objective
functions~\cite{AguTan2007ejor}. Feasible solutions are binary
strings of size~$n$.  The parameter $k$ refers to the number of
variables that influence a particular position from the bit-string
(the epistatic interactions).  The objective function vector is
defined as $f\colon \lbrace 0, 1 \rbrace^{n} \to [0,1)^m$.  Each
objective function is to be maximized, and can be formalized as
follows:
$f_i(x) = \frac{1}{n} \sum_{j=1}^{n} c^i_{j}(x_j, x_{j_1}, \dotsc, x_{j_k})$, $i \in \set{1, \dotsc, m}$,
where $c^i_j \colon \lbrace 0, 1 \rbrace^{k+1} \to [0,1)$ defines
the component function associated with each variable~$x_j$, $j \in \set{1,\dotsc,n}$,
for objective $f_i$, and where $k < n$.
By increasing the number of variable interactions $k$ from $0$
to~$(n-1)$, \RMNKs can be gradually tuned from smooth to rugged.  

We generate an instance of a \RMNK by randomly setting the position of
these epistatic interactions, following a uniform distribution.  The
same epistatic degree $k$
and the same epistatic interactions are used for all the
objectives. 
Component function values are sampled within the range~$[0,
1)$ following a multivariate uniform distribution of dimension~$m$ with a correlation coefficient~$\rho$. 
A positive (resp. negative) correlation coefficient decreases
(resp. increases) the degree of conflict between the objective
function values.

\section{Experimental Setup}
\label{sec:exp}

In the following, we investigate \RMNKs
with a problem size $n\in \set{8,16}$,
an epistatic degree $k \in \set{1, 2, 4, 8}$ such that $k < n$, 
an objective space dimension $m \in \set{2,3,5}$,
and an objective correlation $\rho \in \set{-0.7,-0.2,0.0,0.2,0.7}$ such that $\rho > \frac{-1}{m-1}$, because the corresponding correlation matrix is symmetric positive-definite \cite{VerLieJou2013ejor}.
The investigated problem sizes allow us to enumerate the solution space exhaustively,
and then to solve all the instances to optimality.
One independent random instance is considered for each parameter combination: $\langle \rho, m, n, k \rangle$.
This leads to a total of $91$ problem instances.

In our implementation of PLS for \RMNKs, the neighborhood structure is
taken as the \emph{1-bit-flip}.
The archive is initialized with one random solution from the solution
space.  At each iteration, the neighborhood of the selected solution
is explored exhaustively in a random order.  This order has an impact
on the dynamics of PLS since bounded archiving methods treat incoming
solutions sequentially.  The cost of each iteration of PLS is exactly
$n$ evaluations, corresponding to the neighborhood size.
We experiment with three variants of PLS. \PLSunb corresponds to the
classical PLS with an unbounded archive. \PLShva and \PLSmga use HVA
and MGA, respectively, to bound the size of the archive to a maximum of $\mu$
solutions, where $\mu \in \set{10,20,40,80}$.
We consider $25$ different seeds for the random number generator used in PLS, and we run each PLS variant on each problem instance using each random seed.
This leads to a total of 20$\,$475 runs.

\section{Experimental Analysis}
\label{sec:res}

\subsection{Cardinality of Pareto Local Optimal Sets}

\begin{figure}[!t]
  \centering
\subfloat[\label{rmnkPLSunb/plo-sets-sizes_K_M_N16_r-0.2}$n=16$, $\rho=-0.2$]{\includegraphics[width=0.45\textwidth]{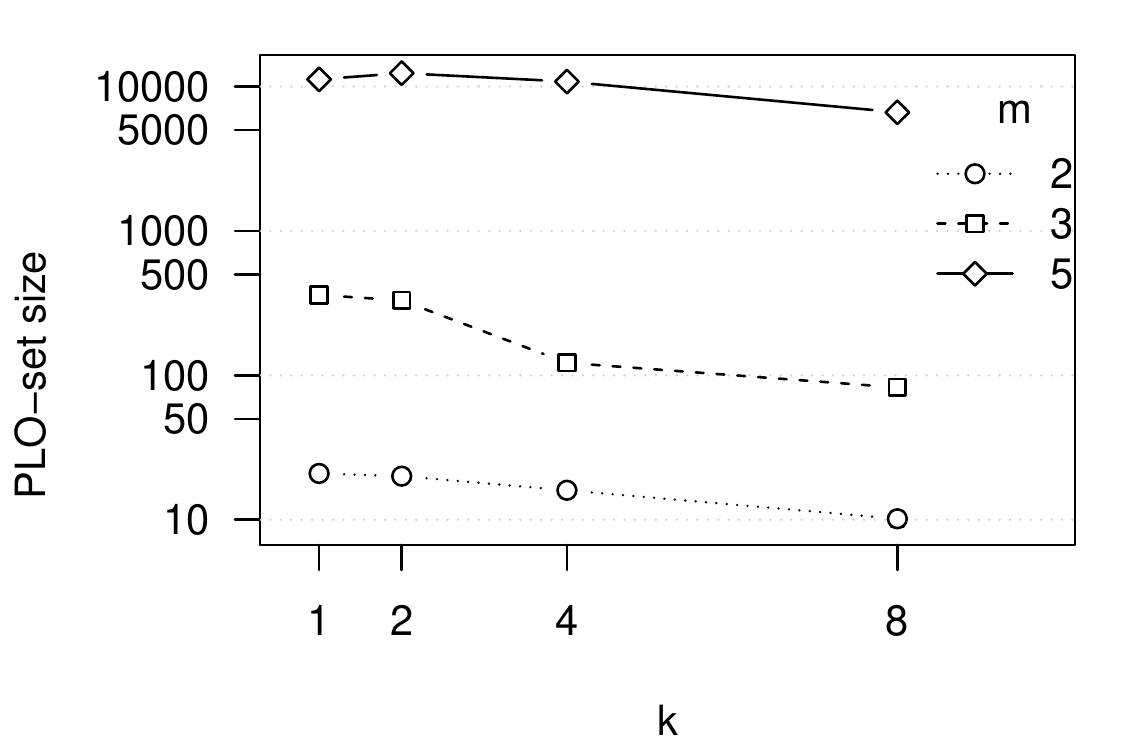}}
\subfloat[$n=16$, $\rho=0.7$]{\includegraphics[width=0.45\textwidth]{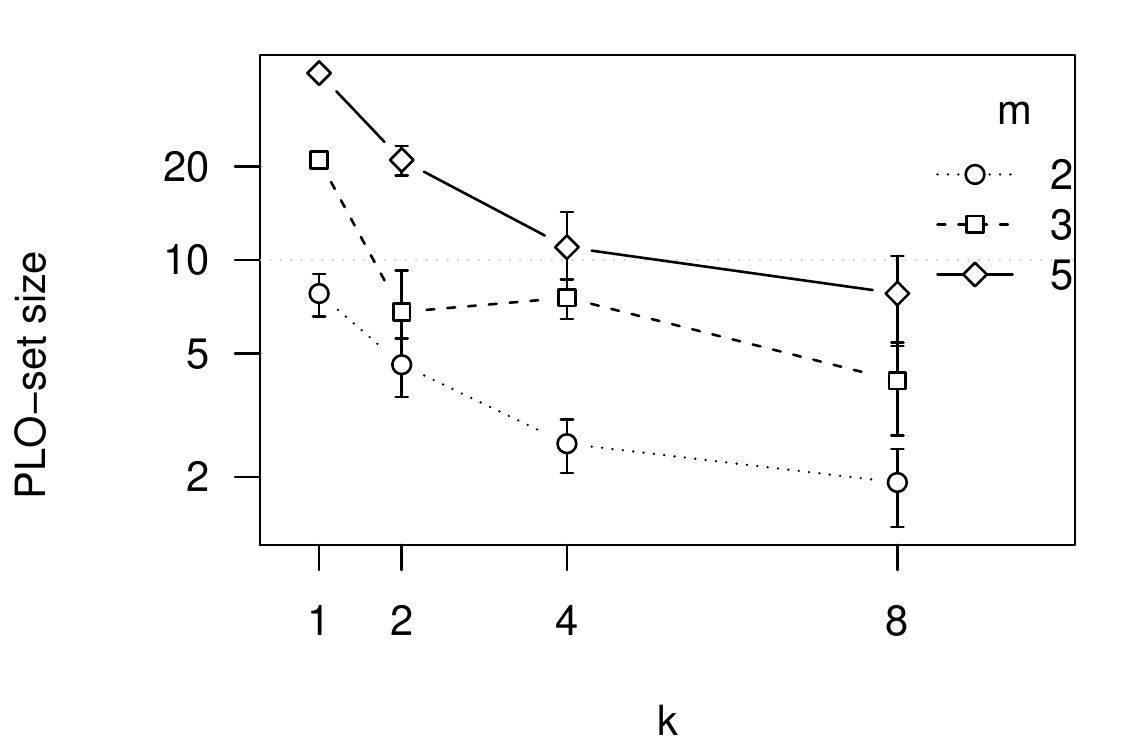}}
\\
\subfloat[\label{rmnkPLSunb/plo-sets-sizes_r_M_N16_K4}$n=16$, $k=4$]{
  \includegraphics[width=0.45\textwidth]{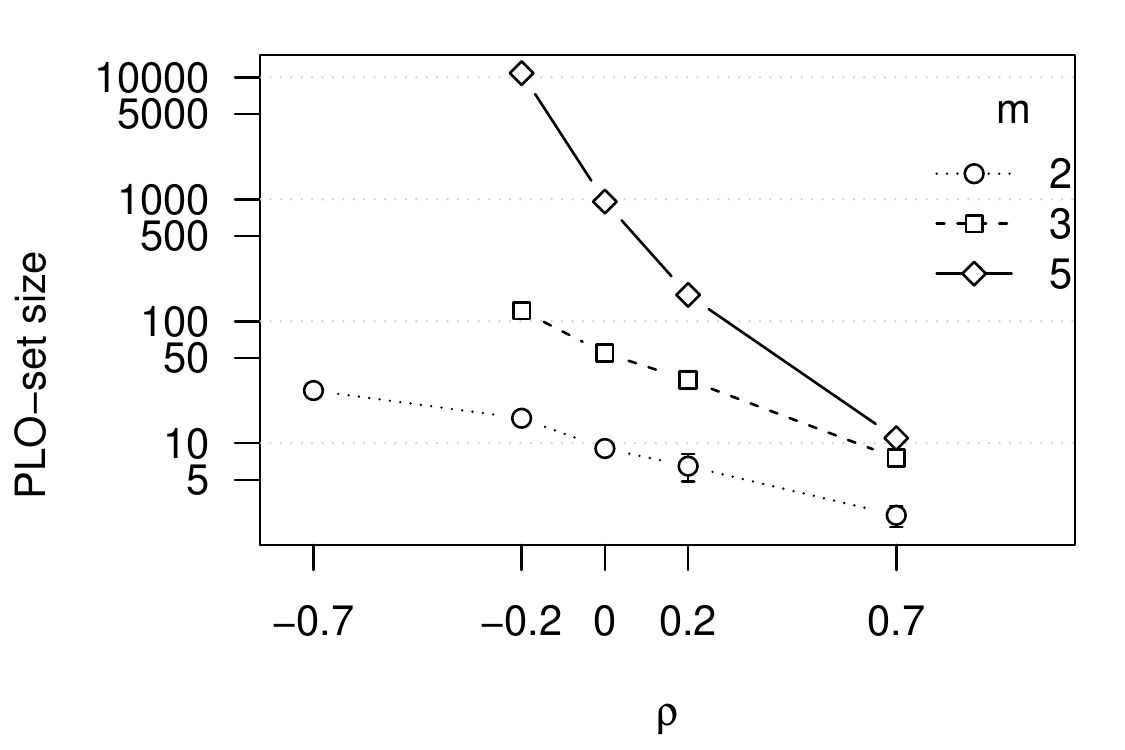}}
\subfloat[\label{rmnkPLSunb/plo-sets-sizes_r_K_N16_M5}$n=16$, $m=5$]{
  \includegraphics[width=0.45\textwidth]{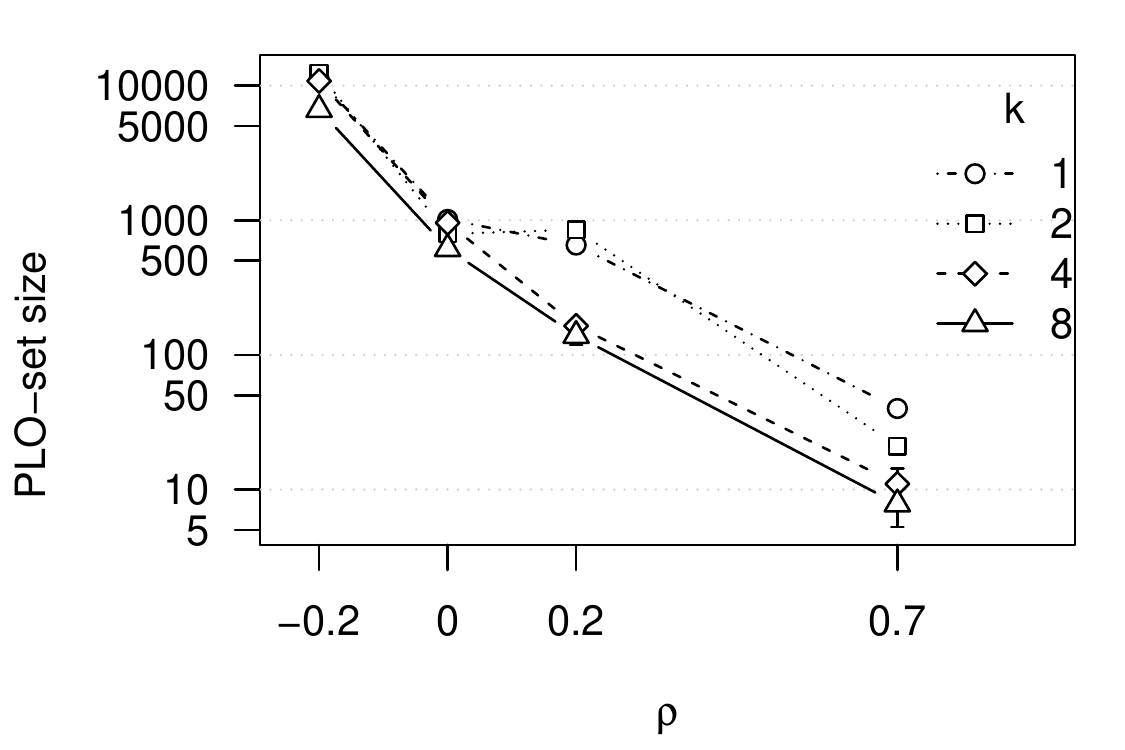}}
\caption{Mean size of the PLO-sets returned by \PLSunb. Error bars give the standard deviation.\label{fig:PLSunb:sizes}}
\end{figure}

Figure~\ref{fig:PLSunb:sizes} shows the size of the PLO-sets identified
by \PLSunb with respect to different instance characteristics. Each
point gives the mean value over the $25$ random seeds on the same
instance, and the error bars indicate the standard deviation. PLO-set
sizes are plotted in logarithmic scale.

In general, the cardinality of PLO-sets increases exponentially with the number
of objectives $m$, as shown in
Figs.~\ref{rmnkPLSunb/plo-sets-sizes_K_M_N16_r-0.2}--\ref{rmnkPLSunb/plo-sets-sizes_r_M_N16_K4}. The correlation between
objective values~$\rho$ is also a crucial factor: the cardinality of the PLO-sets increases exponentially with the linear decrease of $\rho$ (Fig~\ref{rmnkPLSunb/plo-sets-sizes_r_M_N16_K4}). The cardinality
of PLO-sets also increases with lower $k$, but much less noticeably
(Fig.~\ref{rmnkPLSunb/plo-sets-sizes_K_M_N16_r-0.2}).
Moreover, the small error bars in Figs.~\ref{rmnkPLSunb/plo-sets-sizes_K_M_N16_r-0.2}--\ref{rmnkPLSunb/plo-sets-sizes_r_K_N16_M5} indicate that, for a given instance, maximal PLO-sets consistently have roughly the same size.

Given the above results, restricting the size of the PLO-sets by using
bounding archiving methods must have a stronger effect as the
correlation decreases and the number of objectives increases. In the
next section, we examine this effect for each archiving
method.

\subsection{Quality of Local Optimal Sets}

\begin{figure}[!t]
  \centering
\subfloat[\label{all/eps_a_algo_M5_N16_K8_r-0.2}$n=16$, $m=5$, $k=8$, $\rho=-0.2$]{\includegraphics[width=0.45\textwidth]{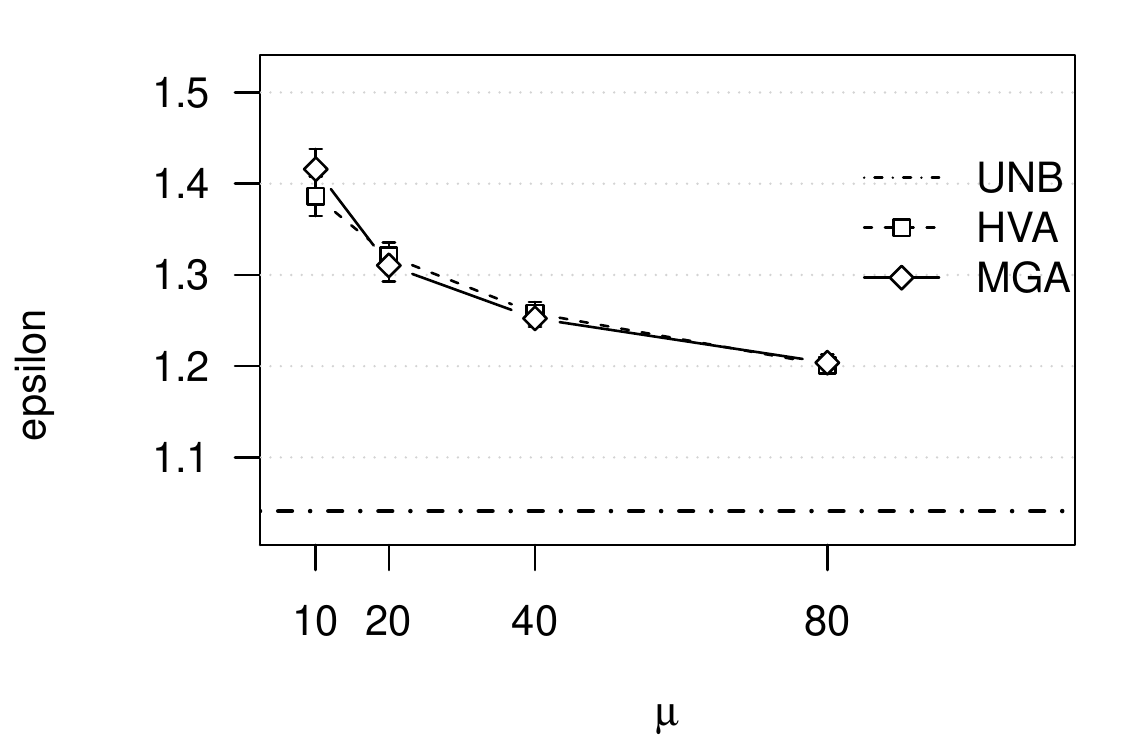}}
\subfloat[\label{all/eps_a_algo_M5_N16_K8_r0}$n=16$, $m=5$, $k=8$, $\rho=0$]{\includegraphics[width=0.45\textwidth]{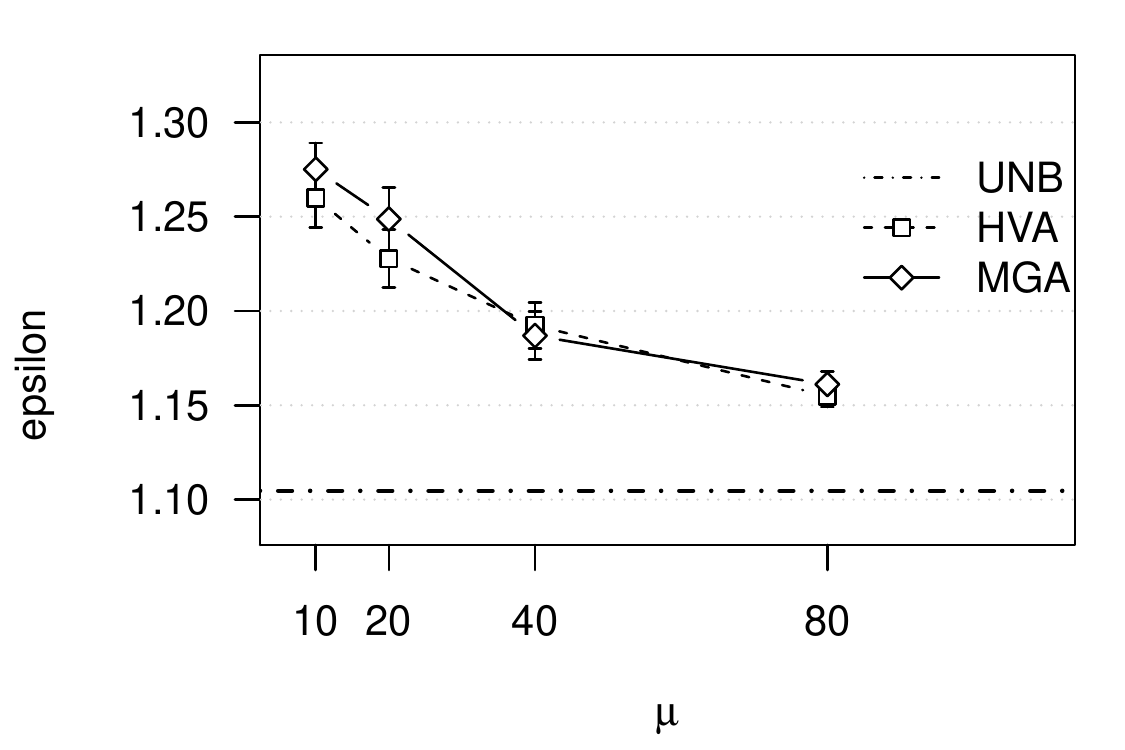}}
\\
\subfloat[\label{all/hv_a_algo_M5_N16_K8_r-0.2}$n=16$, $m=5$, $k=8$, $\rho=-0.2$]{\includegraphics[width=0.45\textwidth]{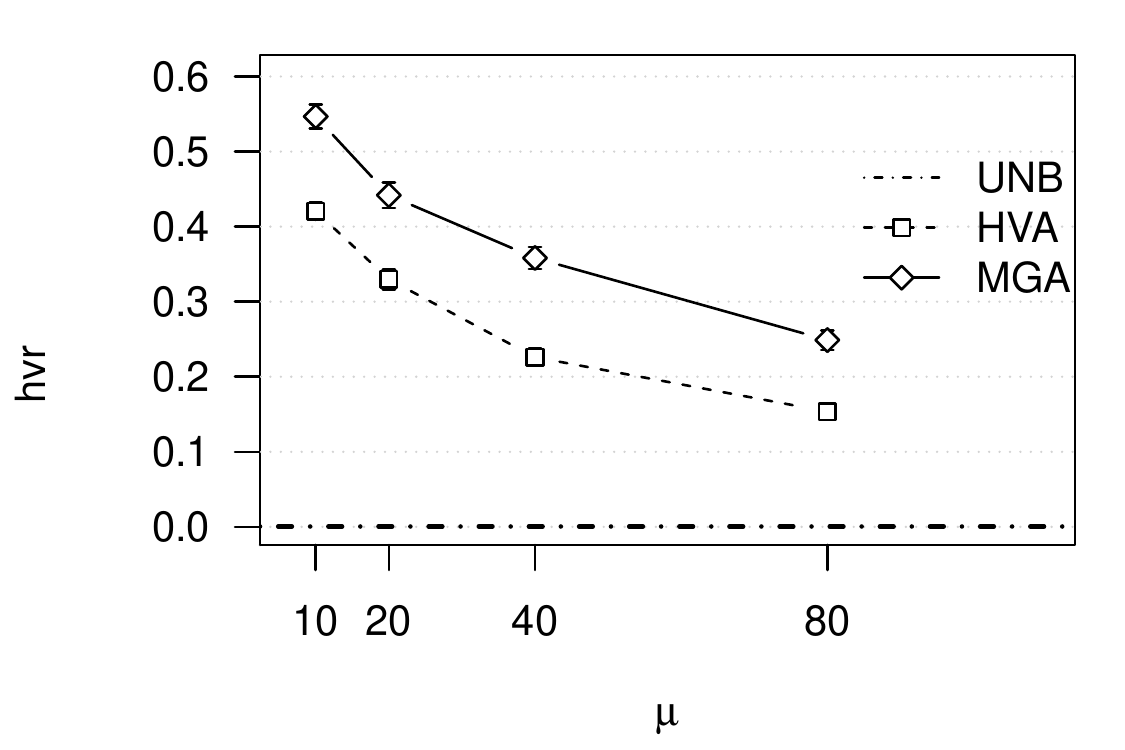}}
\subfloat[\label{all/hv_a_algo_M5_N16_K8_r0}$n=16$, $m=5$, $k=8$, $\rho=0$]{\includegraphics[width=0.45\textwidth]{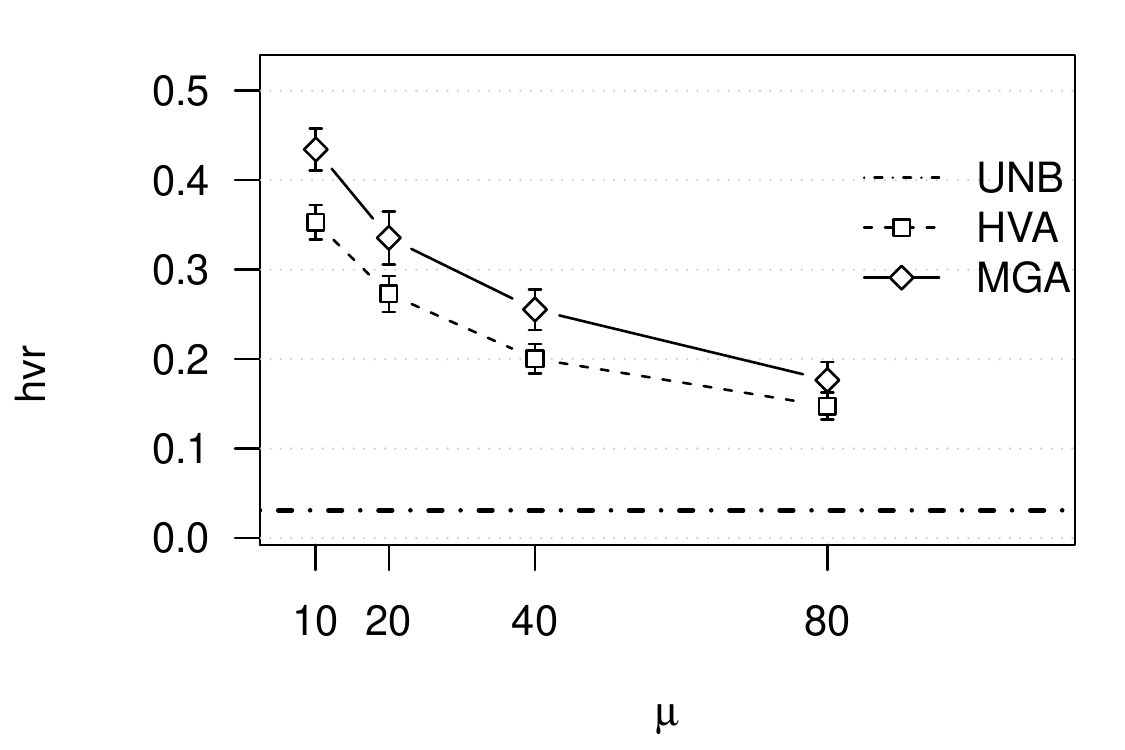}}
\\
\subfloat[\label{all/hv_K_algo_M3_N16_r0_a10}$n=16$, $m=3$, $\rho=0.0, \mu=10$]{\includegraphics[width=0.45\textwidth]{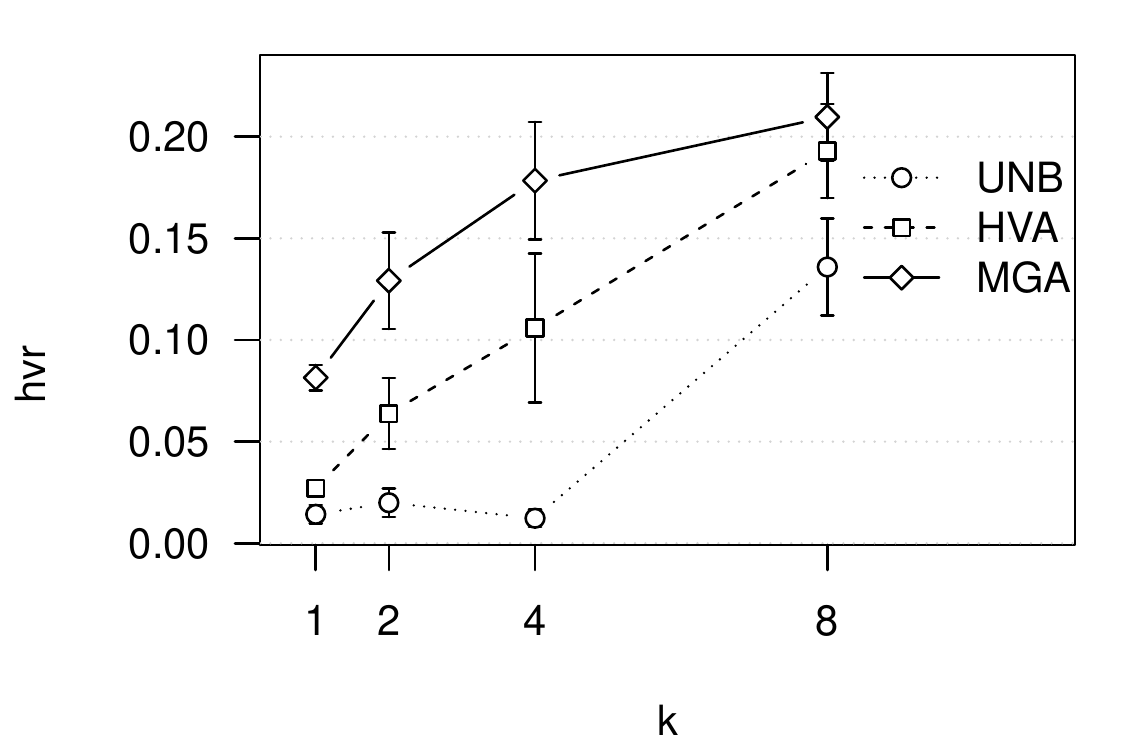}}
\subfloat[\label{all/hv_K_algo_M5_N16_r0_a10}$n=16$, $m=5$, $\rho=0.0, \mu=10$]{\includegraphics[width=0.45\textwidth]{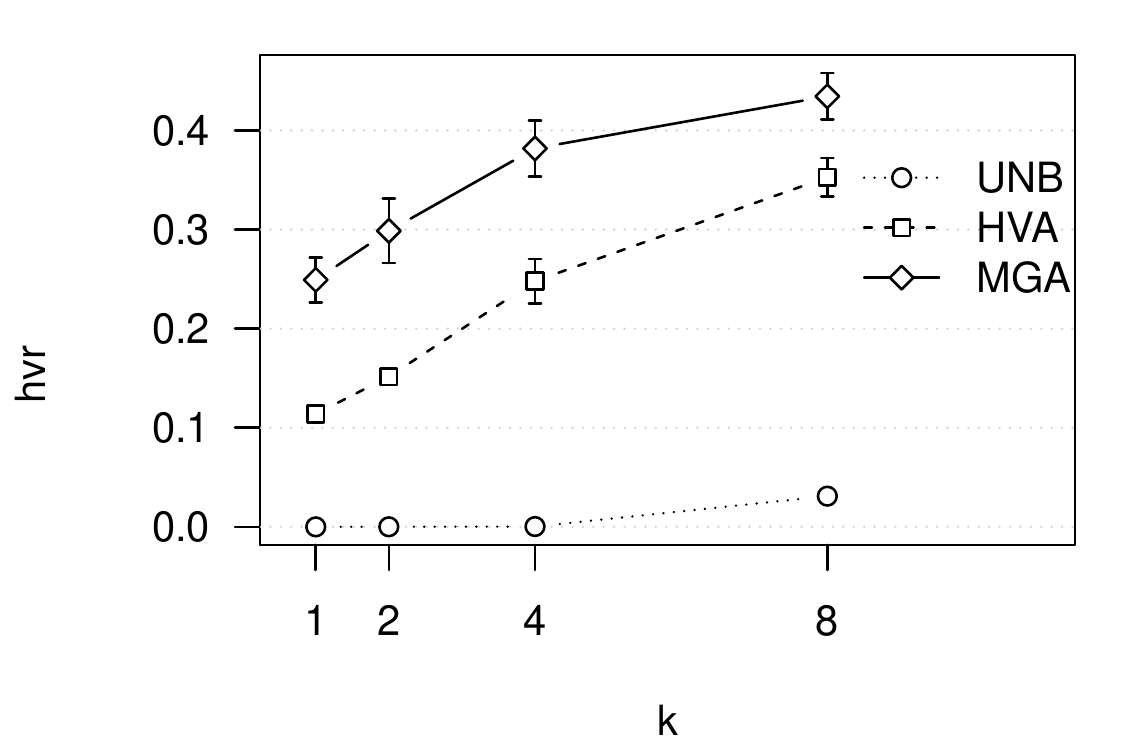}}\\
\caption{Mean quality of the PLO-sets found by \PLSunb, \PLShva and \PLSmga
  measured in terms of multiplicative epsilon (\emph{epsilon}) and
  hypervolume relative difference (\emph{hvr}). Error bars give the standard deviation.\label{fig:quality}}
\end{figure}

We examine the quality of the PLO-sets found by \PLSunb, \PLShva and
\PLSmga in terms of two unary quality measures, namely, the
hypervolume and the multiplicative
epsilon~\cite{ZitThiLauFon2003:tec}.  In order to compare the
hypervolume value for instances with very different characteristics, we
compute the hypervolume relative difference as $hvr(A) = (hv(P)
  - hv(A)) / hv(P)$, where $A \subseteq \OS$ is the image of a PLO-set in
the objective space and $P$ is the exact Pareto front for the instance
under consideration. The reference point is set to the origin.  The
epsilon measure gives the minimum multiplicative factor by which a
PLO-set has to be shifted in the objective space to weakly dominate
the exact Pareto front. Thus, for both measures, a lower value is preferred.
Results are reported in Fig.~\ref{fig:quality} for different parameter settings.

As conjectured above, a first observation is that,  as the size of maximal
PLO-sets increases,  the quality of PLO-sets obtained by bounded
archiving decreases. The increase in quality is almost logarithmic with
respect to the archive size limit ($\mu$), as shown in Figs.~%
\ref{all/eps_a_algo_M5_N16_K8_r-0.2}--\ref{all/hv_a_algo_M5_N16_K8_r0}.
Here we show only results for
$m=5$, but the trends are similar for a lower number of objectives,
although less pronounced. This trend appears independently of whether
we measure quality in terms of epsilon or hypervolume.
Moreover, as in the single-objective case, the average quality of
local optima decreases with $k$ \cite{Kau1993order}, as shown in Figs.~%
\ref{all/hv_K_algo_M3_N16_r0_a10}--\ref{all/hv_K_algo_M5_N16_r0_a10}.

Lastly, there is not much difference in terms of
quality between \PLShva and \PLSmga. When differences occur, the
PLO-sets returned by \PLShva have a better hypervolume (lower
$hvr$ value) than those returned by \PLSmga
(Figs.~\ref{all/hv_a_algo_M5_N16_K8_r-0.2}--\ref{all/hv_a_algo_M5_N16_K8_r0}),
however, there is no clear winner in terms of epsilon value
(Fig.~\ref{all/eps_a_algo_M5_N16_K8_r-0.2}--\ref{all/eps_a_algo_M5_N16_K8_r0}).

\subsection{Difficulty of Identifying Local Optimal Sets}

For single-objective NK-landscapes, the number of iterations, or steps,
of a conventional hill-climbing algorithm provides an estimation of
the average diameter of the basins of attraction of local
optima~\cite{Kau1993order}. This diameter characterizes a problem instance in
terms of multimodality: The larger the length, the larger the basin
diameter and the lower the number of local optima. Conversely, the
smaller the length, the smaller the basin diameter and the higher the
number of local optima. Multimodality characterizes an important
aspect of instance difficulty (\ie, the number of local optima).

\begin{figure}[!t]
  \centering
\subfloat[\label{all/length_a_algo_M3_N16_K8_r-0.2}$n=16$, $m=3$, $k=8$, $\rho=-0.2$]{\includegraphics[width=0.45\textwidth]{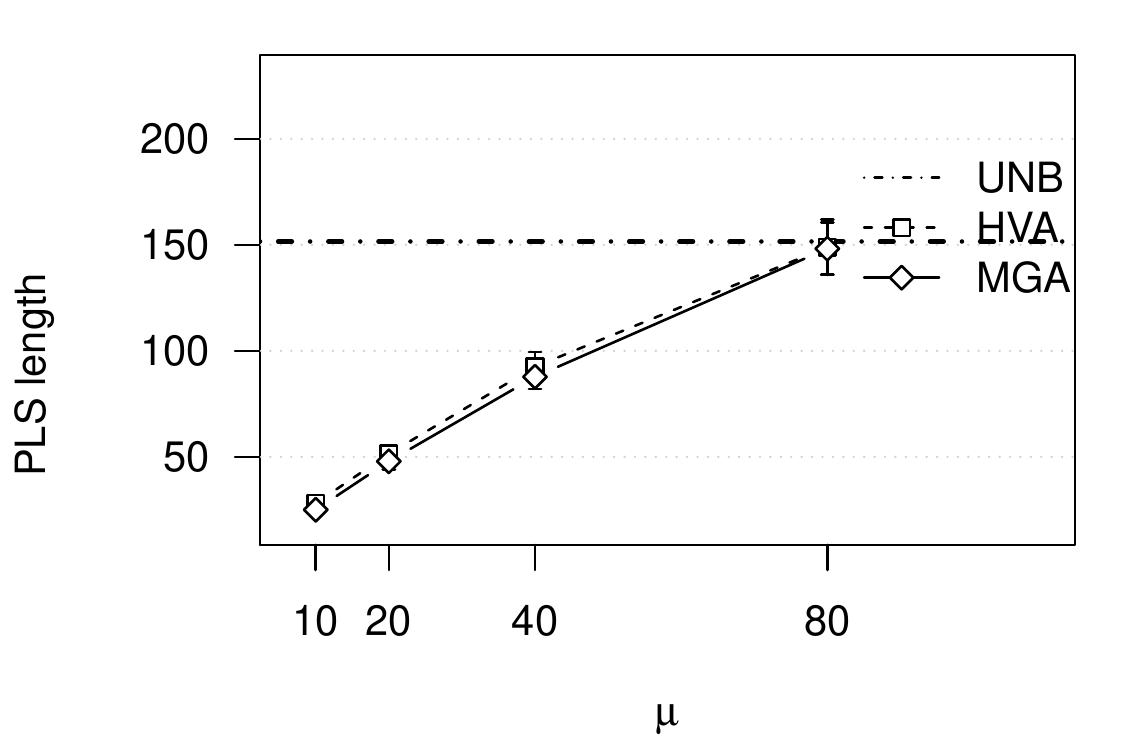}}
\subfloat[\label{all/length_a_algo_M3_N16_K8_r0}$n=16$, $m=3$, $k=8$, $\rho=0$]{\includegraphics[width=0.45\textwidth]{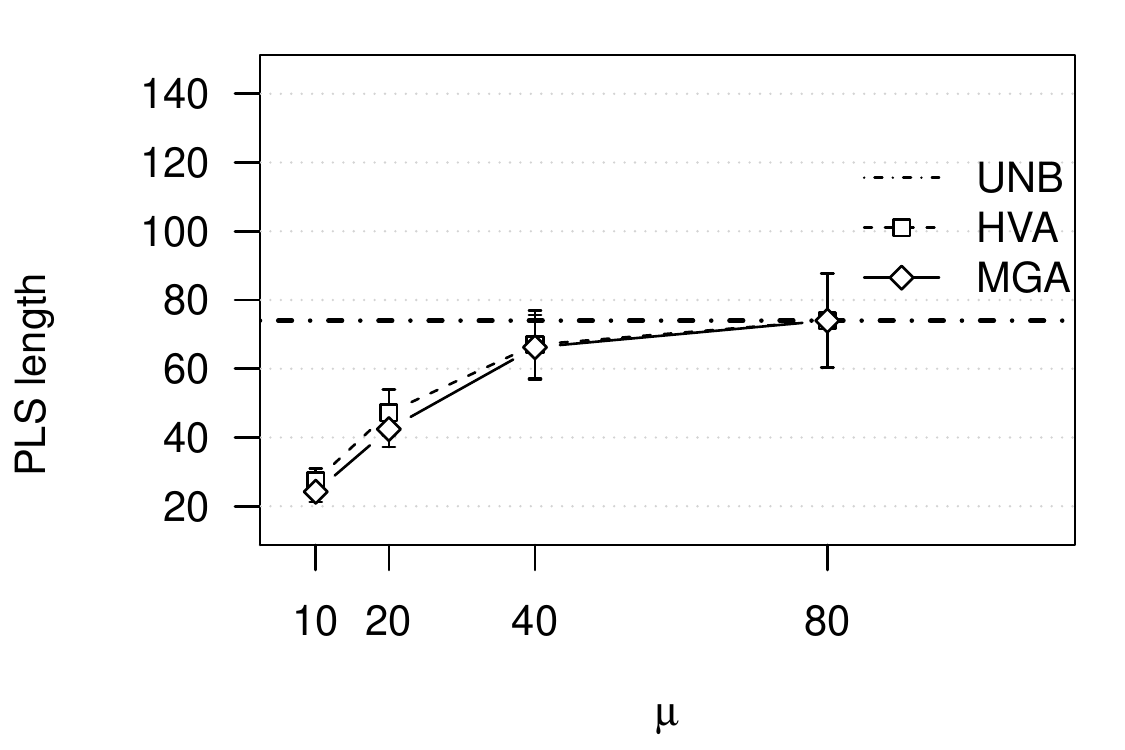}}
\\
\subfloat[\label{all/length_a_algo_M5_N16_K8_r-0.2}$n=16$, $m=5$, $k=8$, $\rho=-0.2$]{\includegraphics[width=0.45\textwidth]{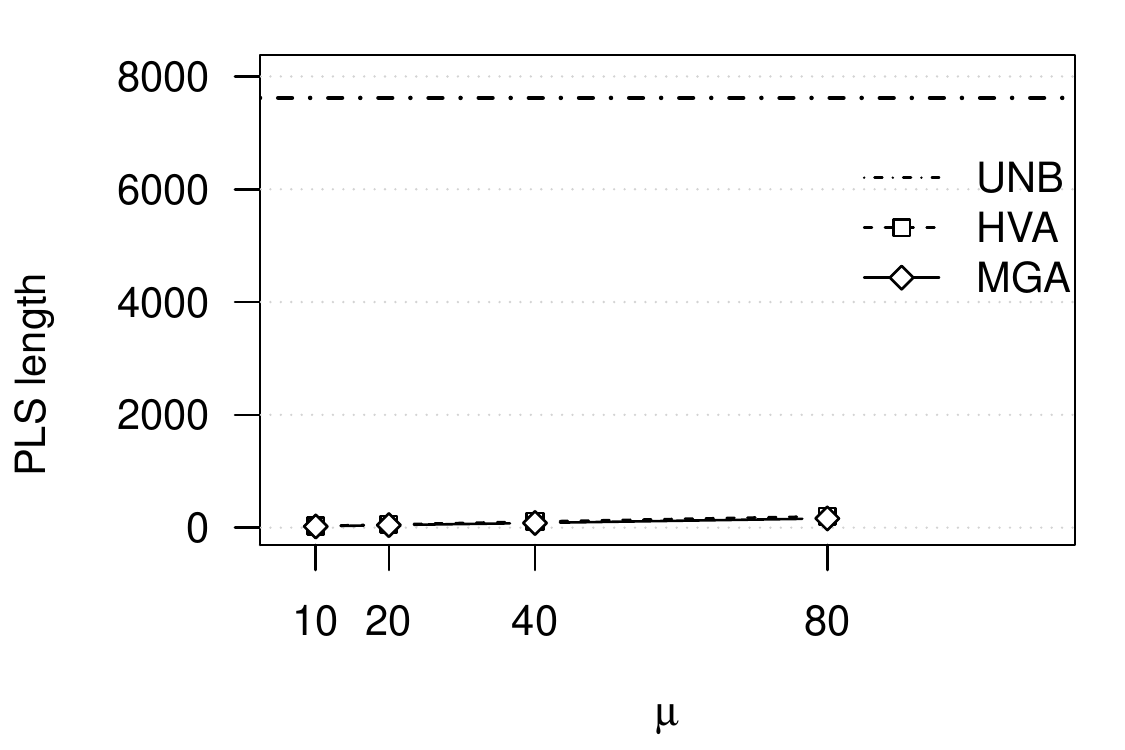}}
\subfloat[\label{all/length_a_algo_M5_N16_K8_r0}$n=16$, $m=5$, $k=8$, $\rho=0$]{\includegraphics[width=0.45\textwidth]{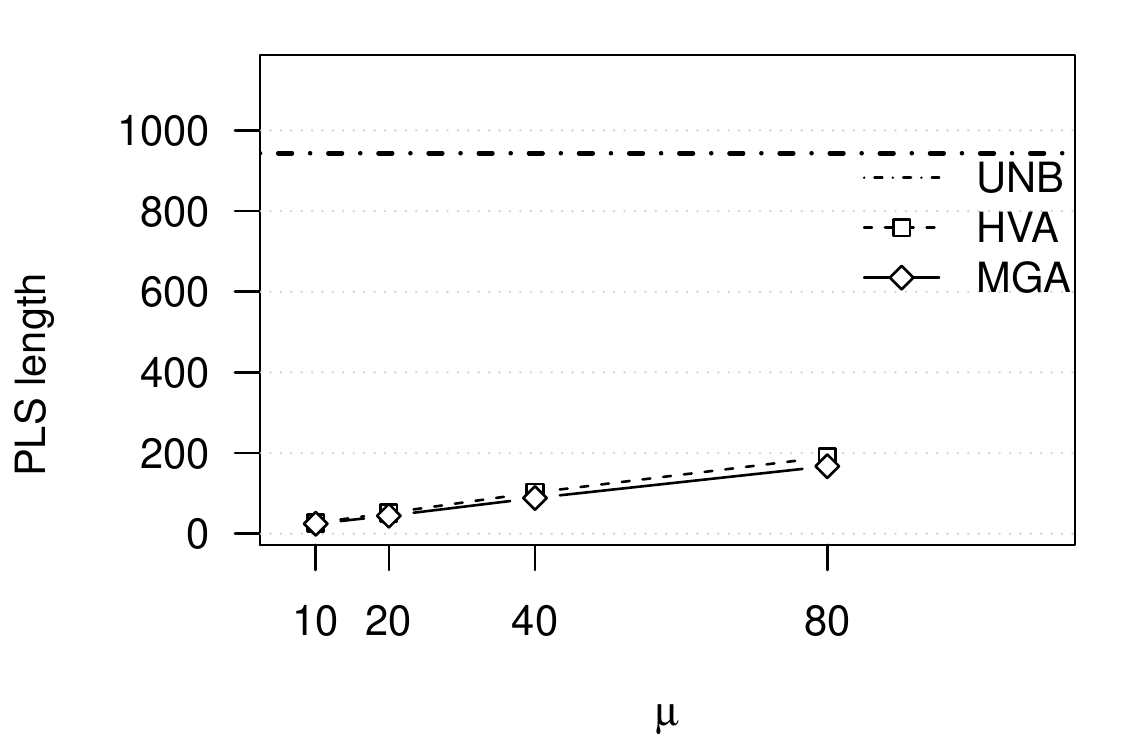}}
\\
\subfloat[\label{all/length_K_algo_M3_N16_r0_a10}$n=16$, $m=3$, $\rho=0.0, \mu=10$]{\includegraphics[width=0.45\textwidth]{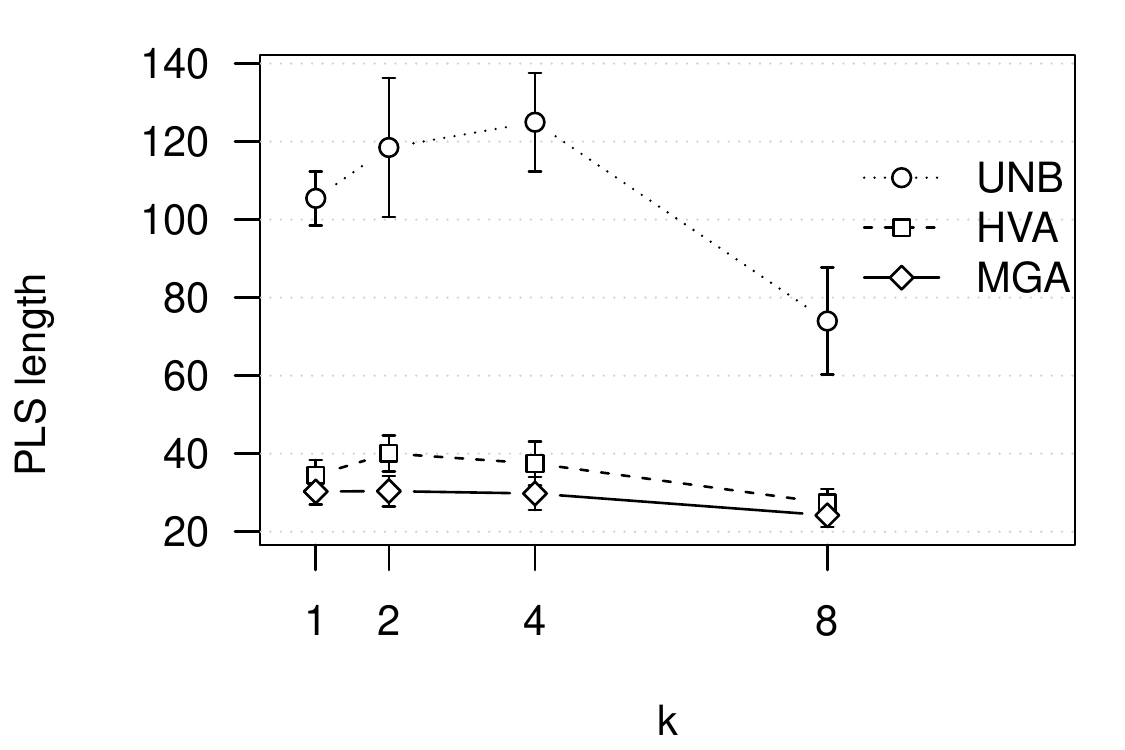}}
\subfloat[\label{all/length_K_algo_M5_N16_r0_a10}$n=16$, $m=5$, $\rho=0.0, \mu=10$]{\includegraphics[width=0.45\textwidth]{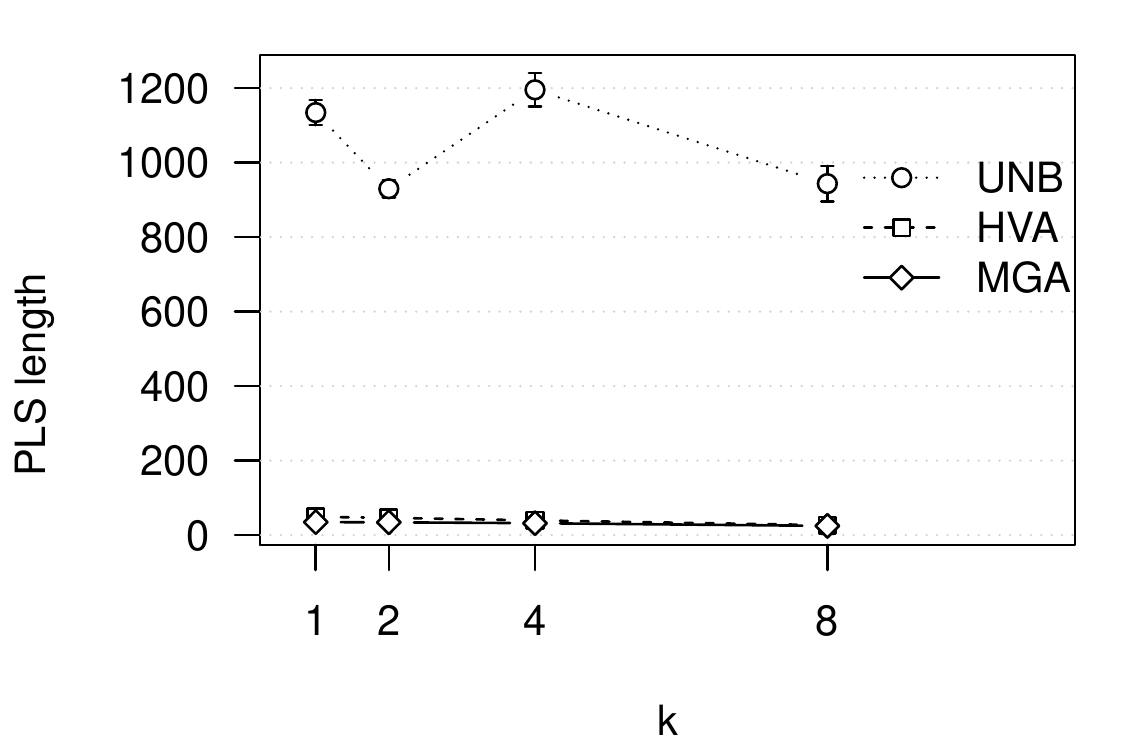}}\\

\caption{Mean length of \PLSunb, \PLShva and \PLSmga. Error bars give the standard deviation.\label{fig:length}}
\end{figure}

As in the single-objective case, the construction of \RMNKs
implies that they are isotropic, that the neighborhood
has the same properties in every direction of the objective space, and
that the basins of attraction have a ball-like shape.
In this section, we report the length of PLS for different archiving
strategies. This allows us to study the running time, in terms of
number of iterations,
required by PLS to identify a PLO-set. 
Moreover, when the PLS length is smaller under one setting than another,
we can reasonably assume that there exist more PLO-sets in the corresponding landscape since the search process got stuck more easily on a
local optima.
Figure~\ref{fig:length} shows the PLS length for the same problem
instances shown in Fig.~\ref{fig:quality}. The comparison between each
pair of plots shows clearly a relationship between the length of PLS
(Fig.~\ref{fig:length}) and the quality of the results
(Fig.~\ref{fig:quality}): Larger length corresponds to better quality
in general.
Interestingly, bounding the archive size substantially reduces both the quality of
the obtained approximation as well as the running time of PLS. The smaller the archive
size, the larger the difference with \PLSunb.
However, this also has the effect of increasing the number of PLO-sets.
Indeed, when the archive size is small, the PLS length is small, which  suggests that 
the average distance between a local optimum and the solutions from the corresponding basin of attraction
is also small and, hence, the number of PLO-sets is large.

Surprisingly, when the archive is unbounded as reported in
Fig. \ref{fig:PLSunb:sizes},
the PLS length increases from $k=1$ to $k=4$ (despite we know that the number of PLO-solutions increases linearly with $k$~\cite{VerLieJou2013ejor}) and only becomes smaller for $k=8$ (Fig.~\ref{fig:length}).
This increase in PLS length contradicts  known results from single-objective optimization~\cite{Kau1993order}.
In fact, in the case of PLO-sets, both the number of PLO-solutions and the average size of the neighborhood of a PLO-set influence the number of PLO-sets.
First, when the number of PLO-solutions increases, 
it is expected that the number of PLO-sets also increases, which will decrease the PLS length.
By contrast, when a PLO-set is larger, the number of neighbor solutions in this set is larger as well.
This potentially reduces the number of PLO-sets, making PLS run longer.
The balance between these two effects could explain the PLS length;
\ie, the PLS length starts to decrease when the number of PLO-solutions has a larger impact on the number of PLO-sets than the size of PLO-sets.
Overall, these results suggest
that the length of PLS could provide an estimation of the number of
PLO-sets, and thus a measure of difficulty for archive-based
local search.
%

\section{Conclusions}

In this paper, we analyzed the characteristics of local optima in set-based multi-objective local search when applied to multi-objective NK-landscapes with correlated objectives. 
First, the main factors affecting the cardinality of the maximal PLO-sets
returned by \PLSunb are the number of objectives and the correlation
between them. By changing these two factors, the PLO-set size can
vary from a few tens to tens of thousands. Our results confirm trends
already noticed for other MCOP problems. In particular, the
exponential increase in the PLO-set size with lower objective correlation
has already been reported for the bi-objective QAP~\cite{PaqStu06:mqap}.
Another interesting observation is that, given a particular instance,
the variability of PLO-set sizes is usually a very small fraction of
the average size, that is, most maximal PLO-sets for a given instance
have roughly the same size. This is not an obvious conjecture to make,
and we currently do not know if it is also the case for other MCOPs.
Our experiments also strongly indicate that the relationship between
local search length and number of local optima, which is well-studied
in the single-objective case~\cite{Kau1993order} and in the case of
PLO-solutions~\cite{VerLieJou2013ejor},
also applies to PLS and PLO-sets. Our results clearly show that shorter 
PLS lengths typically correspond to lower quality results (and hence,
more difficult instances). A precise estimation of this relationship
in the case of PLO-sets would require to determine the exact number
of PLO-sets for a given~instance. 

From an algorithm design point of view, this work helps to better capture
the relation between running time and approximation quality according to
the problem instance characteristics.
From a theoretical point of view, it would be interesting to
understand precisely the relationships between the PLO-sets obtained
by each archiving method. Moreover, it is clear to us that there is a
direct relationship between the PLO-solutions of a problem, and the
number and size of PLO-sets, however, a precise formulation remains to
be described. Finally, we left for future work discussing the
implications of the results reported here with respect to
theoretical bounds reported in the
literature~\cite{BriFri2011gecco}.
Finally, complementary studies on other MCOPs and larger
problem instances would allow us to better understand the structure of
PLO-sets for different archiving techniques.

\vspace{-1.0ex}
{\footnotesize
  \paragraph*{Acknowledgments.} 
  This work is one of the result from the discussions at the SIMCO Workshop at the Lorentz-Center. 
  Manuel L\'opez-Ib\'a\~nez acknowledges support from the Belgian
  F.R.S.-FNRS, of which he is a postdoctoral researcher.}

\vspace{-1.7ex}
\bibliographystyle{splncs03abbrev.bst}

\providecommand{\MaxMinAntSystem}{{$\cal MAX$--$\cal MIN$} {A}nt {S}ystem}
  \providecommand{\Rpackage}[1]{#1} \providecommand{\SoftwarePackage}[1]{#1}
  \providecommand{\proglang}[1]{#1}
\begin{thebibliography}{10}
\providecommand{\url}[1]{\texttt{#1}}
\providecommand{\urlprefix}{URL }

\bibitem{AguTan2007ejor}
Aguirre, H.E., Tanaka, K.: Working principles, behavior, and performance of
  {MOEAs} on {MNK}-landscapes. Eur. J. Oper. Res.  181(3),  1670--1690 (2007)

\bibitem{Bor1998}
Borges, P.C., Hansen, M.P.: A basis for future successes in multiobjective
  combinatorial optimization. Tech. Rep. IMM-REP-1998-8, Institute of
  Mathematical Modelling, Technical University of Denmark, Lyngby, Denmark
  (1998)

\bibitem{BriFri2011gecco}
Bringmann, K., Friedrich, T.: Convergence of hypervolume-based archiving
  algorithms~{I}: Effectiveness. In: Krasnogor, N., et~al. (eds.) Proceedings
  of the Genetic and Evolutionary Computation Conference, GECCO 2011, pp.
  745--752. ACM Press, New York, NY (2011)

\bibitem{DruThi2012}
Drugan, M.M., Thierens, D.: Stochastic pareto local search: Pareto
  neighbourhood exploration and perturbation strategies. Journal of Heuristics
  18,  727--766 (2012)

\bibitem{DubLopStu2011cor}
Dubois-Lacoste, J., L{\'o}pez-Ib{\'a}{\~n}ez, M., St{\"u}tzle, T.: A hybrid
  {TP$+$PLS} algorithm for bi-objective flow-shop scheduling problems. Comput.
  Oper. Res.  38(8),  1219--1236 (2011)

\bibitem{MMO2004}
Gandibleux, X., et~al. (eds.): Metaheuristics for Multiobjective Optimisation.
  LNEMS, Springer, Berlin, Germany (2004)

\bibitem{Kau1993order}
Kauffman, S.A.: The Origins of Order. Oxford University Press (1993)

\bibitem{Knowles2002PhD}
Knowles, J.D.: Local-Search and Hybrid Evolutionary Algorithms for {P}areto
  Optimization. Ph.D. thesis, University of Reading, UK (2002)

\bibitem{KnoCor2004lnems}
Knowles, J.D., Corne, D.: Bounded {P}areto archiving: {T}heory and practice.
  In: Gandibleux et~al.  \cite{MMO2004}, pp. 39--64

\bibitem{LauZen2011ejor}
Laumanns, M., Zenklusen, R.: Stochastic convergence of random search methods to
  fixed size {P}areto front approximations. Eur. J. Oper. Res.  213(2),
  414--421 (2011)

\bibitem{LopKnoLau2011emo}
L{\'o}pez-Ib{\'a}{\~n}ez, M., Knowles, J.D., Laumanns, M.: On sequential online
  archiving of objective vectors. In: Takahashi, R.H.C., et~al. (eds.)
  Evolutionary Multi-criterion Optimization (EMO 2011), LNCS, vol. 6576, pp.
  46--60. Springer (2011)

\bibitem{Lust09}
Lust, T., Teghem, J.: Two-phase {P}areto local search for the biobjective
  traveling salesman problem. Journal of Heuristics  16(3),  475--510 (2010)

\bibitem{LusTeg2012itor}
Lust, T., Teghem, J.: The multiobjective multidimensional knapsack problem: a
  survey and a new approach. International Transactions in Operational Research
   19(4),  495--520 (2012)

\bibitem{PaqChiStu2004mmo}
Paquete, L., Chiarandini, M., St{\"u}tzle, T.: {P}areto local optimum sets in
  the biobjective traveling salesman problem: An experimental study. In:
  Gandibleux et~al.  \cite{MMO2004}, pp. 177--200

\bibitem{PaqSchStu07:aor}
Paquete, L., Schiavinotto, T., St{\"u}tzle, T.: On local optima in
  multiobjective combinatorial optimization problems. Annals of Operations
  Research  156,  83--97 (2007)

\bibitem{PaqStu06:mqap}
Paquete, L., St{\"u}tzle, T.: A study of stochastic local search algorithms for
  the biobjective {QAP} with correlated flow matrices. Eur. J. Oper. Res.
  169(3),  943--959 (2006)

\bibitem{VerLieJou2013ejor}
Verel, S., Liefooghe, A., Jourdan, L., Dhaenens, C.: On the structure of
  multiobjective combinatorial search space: {MNK}-landscapes with correlated
  objectives. Eur. J. Oper. Res.  227(2),  331--342 (2013)

\bibitem{ZitThiLauFon2003:tec}
Zitzler, E., Thiele, L., Laumanns, M., Fonseca, C.M., {Grunert da Fonseca}, V.:
  Performance assessment of multiobjective optimizers: an analysis and review.
  {IEEE} Trans. Evol. Comput.  7(2),  117--132 (2003)

\end{thebibliography}
\providecommand{\MaxMinAntSystem}{{$\cal MAX$--$\cal MIN$} {A}nt {S}ystem}
  \providecommand{\Rpackage}[1]{#1} \providecommand{\SoftwarePackage}[1]{#1}
  \providecommand{\proglang}[1]{#1}

\end{document}